%% file: main.tex
\documentclass[10pt,twocolumn,letterpaper]{article}

\usepackage[pagenumbers]{cvpr} 

\input{preamble}

\definecolor{cvprblue}{rgb}{0.21,0.49,0.74}
\usepackage[pagebackref,breaklinks,colorlinks,citecolor=cvprblue]{hyperref}

\newcommand*{\method}{SYM3D}
\usepackage{pifont}
\usepackage{xspace}
\usepackage{multirow}
\usepackage{textcomp}
\usepackage{amsmath}
\usepackage{caption}
\usepackage{subcaption}
\usepackage{adjustbox}
\usepackage{wrapfig}
\newcommand{\cmark}{\ding{51}}%
\newcommand{\xmark}{\ding{55}}%

\makeatletter
\DeclareRobustCommand\onedot{\futurelet\@let@token\@onedot}
\def\@onedot{\ifx\@let@token.\else.\null\fi\xspace}
 
\def\ie{{\em i.e}\onedot}

\title{SYM3D: Learning Symmetric Triplanes for Better 3D-Awareness of GANs}

\author{
  Jing Yang, Kyle Fogarty, Fangcheng Zhong, Cengiz Oztireli\\
  University of Cambridge, UK\\
}

\begin{document}
\maketitle
\input{sec/0_abstract}

\input{sec/1_intro}
\input{sec/2_related}
\input{sec/3_method}
\input{sec/4_exp}
\input{sec/5_conclusion}

{
    \small
    \bibliographystyle{ieeenat_fullname}
    \bibliography{main}
}

\end{document}

%% file: preamble.tex
\usepackage[dvipsnames]{xcolor}


%% file: sec/0_abstract.tex
\begin{abstract}

Despite the growing success of 3D-aware GANs, which can be trained on 2D images to generate high-quality 3D assets, they still rely on multi-view images with camera annotations to synthesize sufficient details from all viewing directions. 
However, the scarce availability of calibrated multi-view image datasets, especially in comparison to single-view images, has limited the potential of 3D GANs.
Moreover, while bypassing camera pose annotations with a camera distribution constraint reduces dependence on exact camera parameters, it still struggles to generate a consistent orientation of 3D assets.
To this end, we propose \method{}, a novel 3D-aware GAN designed to leverage the prevalent reflectional symmetry structure found in natural and man-made objects, alongside a proposed view-aware spatial attention mechanism in learning the 3D representation.
We evaluate \method{} on both synthetic (ShapeNet Chairs, Cars, and Airplanes) and real-world datasets (ABO-Chair), demonstrating its superior performance in capturing detailed geometry and texture, even when trained on only single-view images. 
Finally, we demonstrate the effectiveness of incorporating symmetry regularization in helping reduce artifacts in the modeling of 3D assets in the text-to-3D task. Project is at \url{https://jingyang2017.github.io/sym3d.github.io/}
\end{abstract}

%% file: sec/1_intro.tex
\vspace{-4mm}
\section{Introduction} \label{sec:intro}

Efficient 3D asset generation is crucial for a multitude of industries such as video games, film-making, digital fabrication, design, and VR/AR. However, collecting large-scale, high-quality 3D datasets is difficult, hindering the training of generative models for high-fidelity 3D assets. Advances in neural implicit representations and differentiable rendering have led to the success of 3D-aware GANs that can be trained on 2D images~\cite{pi-gan,schwarz2020graf,eg3d}. With a differentiable renderer, these methods enable 3D asset generation with guidance from an image discriminator. This significantly enhances data accessibility by bypassing the need for ground-truth 3D assets during training. 

Despite their impressive performance, current 3D GANs face several challenges.
Multi-view images remain indispensable for synthesizing 3D assets with plausible details from all viewing angles~\cite{get3d,shue20233d,ssdnerf}. 
While it is possible to train a 3D GAN using only single-view images, the quality declines substantially. 
Meanwhile, 3D GANs without camera annotations as supervision often produce 3D assets with arbitrary orientations, posing difficulties for users in accurately positioning these assets in downstream applications, due to the absence of a consistent local coordinate frame. 
Furthermore, learning objects in a consistent orientation allows the model to exploit structural information across examples.
Restricting the sampling of camera locations during the rendering stage of training to align with the camera distribution in the training dataset does not fully resolve this issue~\cite{schwarz2020graf,get3d}.
Given the scarcity of multi-view datasets with accurate camera annotations in real-world applications, this necessitates an approach capable of learning from single-image settings, where each object offers only one view for training, with only camera distribution. 

We discovered that the aforementioned bottlenecks of 3D GANs can be simultaneously resolved by exploiting an explicit reflectional symmetry prior, prevalent in both natural and man-made objects~\cite{mitra2013symmetry}, in the 3D representation.
With a symmetry structure, an object can be divided into two halves, with one half approximately mirroring the other.
In nature, numerous organisms exhibit reflectional symmetry (\textit{e.g.}, butterflies with their wings or leaves with their veins). 
Similarly, various artifacts (\textit{e.g.}, buildings, vehicles, and furniture) demonstrate symmetry for aesthetic appeal, structural stability, functional efficiency, and ease of production.
Our insight is that the symmetry prior encodes information about the unseen view of an object when only a single view is accessible. 
Additionally, the symmetry plane implicitly defines a canonical frame shared by various object instances.
Therefore, for these types of objects, employing the symmetry prior not only enhances model performance, but also enforces the generated assets to be positioned under a consistent canonical frame. Most importantly, symmetrical structures are themselves vital properties to adhere to in 3D asset generation.

\input{fig/3_chair_s3_shape_texture}
To this end, we introduce \method{}, a 3D-aware GAN that can be trained on \emph{single-view} images \emph{without} known camera poses to generate 3D textured meshes for objects with symmetrical structures. Following GET3D~\cite{get3d}, we employ two triplane representations~\cite{eg3d} separately for the geometry and texture. Additionally, we integrate two components into the triplane representation: view-wise spatial attention and symmetry regularization. View-wise spatial attention guides each feature plane within the geometry triplane to focus on distinct viewpoints of a scene. Reflectional symmetry encourages the feature planes to respect the natural symmetry~\cite{mitra2013symmetry} of the object being modeled, reducing the ambiguity in learning a consistent orientation for an object category when the ground truth camera pose is unknown. 
Figure~\ref{fig:chair_s3} shows a comparison between GET3D and \method{} in generating 3D assets.
The main contributions of this paper are listed below:
\vspace{-4mm}
\begin{itemize}
\item We introduce a novel symmetry-aware triplane representation for improving 3D awareness of GANs with minimal additional computational cost, effectively capturing both geometry and texture from single images of diverse symmetric objects within a specific category.
\item We are the first to incorporate symmetry regularization into the learning process for both geometry and texture in triplanes, enabling the model to deliver a complete representation of symmetric objects, even when trained on datasets with incomplete views.
\item Our experiments demonstrate the superior performance of \method{} in generating high-quality 3D assets from 2D image collections across a variety of categories of symmetric objects (chair, car, airplane), datasets (synthetic data ShapeNet~\cite{shapenet}, real-world Amazon Berkeley Objects~\cite{collins2022abo}), and training set view distributions (complete 360-degree views and partial 240-degree views). 
Additionally, we demonstrate that symmetry regularization effectively reduces artifacts in the modeling of 3D assets in the text-to-3D task.
\vspace{-2mm}
\end{itemize}

%% file: fig/3_chair_s3_shape_texture.tex
\begin{figure*}[t]
    \vspace{-2mm}
    \centering
    \includegraphics[width=0.96\linewidth]{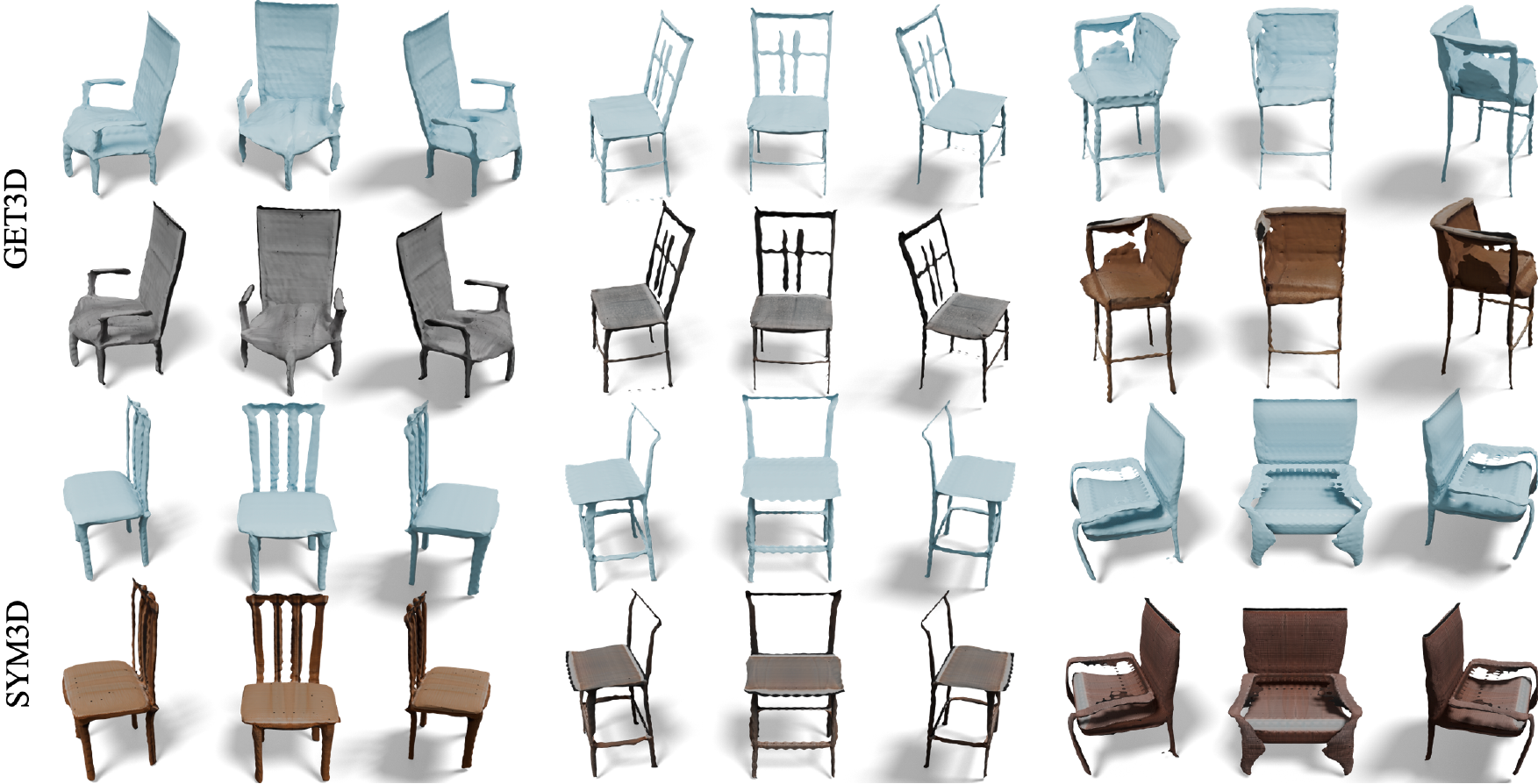}
    \caption{Comparison of shapes generated by GET3D \cite{get3d} and our \method{}, rendered in Blender. \method{} learns symmetric triplanes for improving 3D-awareness of GANs. Compared to GET3D, \method{} can synthesize diverse objects with reasonable geometry and texture after training it on datasets with incomplete views. Refer Section \ref{datasets} for dataset details.}
    \label{fig:chair_s3}
    \vspace{-6mm}
\end{figure*}

%% file: sec/2_related.tex
\section{Related work} \label{sec:related_work}
\vspace{-2mm}

\noindent {\bf 3D-aware GANs with Triplane Representation.}
Since EG3D~\cite{eg3d} introduces the concept of triplane representation, this representation has become increasingly popular in 3D modeling for its efficiency and effectiveness.
A line of works focuses on improving triplane representations,
for example,
He et al.~\cite{orthoplanes} inject an extra axis to 2D triplane by preserving information associated with projection distance.
Wu and Zheng~\cite{wu2022learning} develop a multi-scale triplane in a hierarchical fashion that enables learning 3D shapes from a single example, progressively refining from coarse to detailed features. 
Another line of works distills the triplane representation to describe a single object.
Trevithick et al.~\cite{trevithick2023real} learn a triplane representation to describe an unposed face from a pretrained EG3D, which facilitates real-time, photo-realistic 3D face rendering.
Bhattarai et al.~\cite{bhattarai2024triplanenet} improve GAN inversion by learning offsets to adjust the triplane representation.
Our work improves triplane representation by introducing structure awareness, which enhances the triplane's ability to capture distinct viewpoints of an object. This enhancement is achieved in conjunction with symmetry regularization, allowing for a more holistic representation in training 3D-aware generators. 

\input{fig/1_sym3d_overview}

\noindent {\bf Reflectional Symmetry in Image Synthesis.}
Symmetric objects are widespread in nature, architecture, and art, and reflectional symmetry plays a key role in human visual perception~\cite{mitra2013symmetry}. This concept has been extensively utilized in computer vision research. A significant research direction in this domain focuses on applying reflectional symmetry principles to enhance the rendering of 2D images. 
Wu et al.~\cite{wu2020unsupervised} investigate the inference of 3D deformable objects from single images by employing symmetric structures to distinguish between depth, albedo, viewpoint, and illumination components. 
NeRD~\cite{xu2020ladybird} presents a neural detector designed to identify 3D reflection symmetry in objects, estimating the normal vectors of their mirror planes. 
Yin et al.~\cite{yin20233d} have advanced 3D GAN inversion techniques by training with mirrored images, leveraging symmetry to enhance the quality of the results. 
SymmNeRF~\cite{symmnerf} represents an innovative strategy that explicitly incorporates symmetry into the training of neural radiance fields, utilizing both pixel-aligned image features and their symmetric analogs as additional training inputs.
We introduce symmetry regularization to both geometry and texture, ensuring more accurate and realistic 3D asset generation. Our insight is that the symmetry prior encodes information about the unseen view of an object when only a single view is accessible.


%% file: fig/1_sym3d_overview.tex
\begin{figure*}[t]
    \centering
    \includegraphics[width=0.96\linewidth]{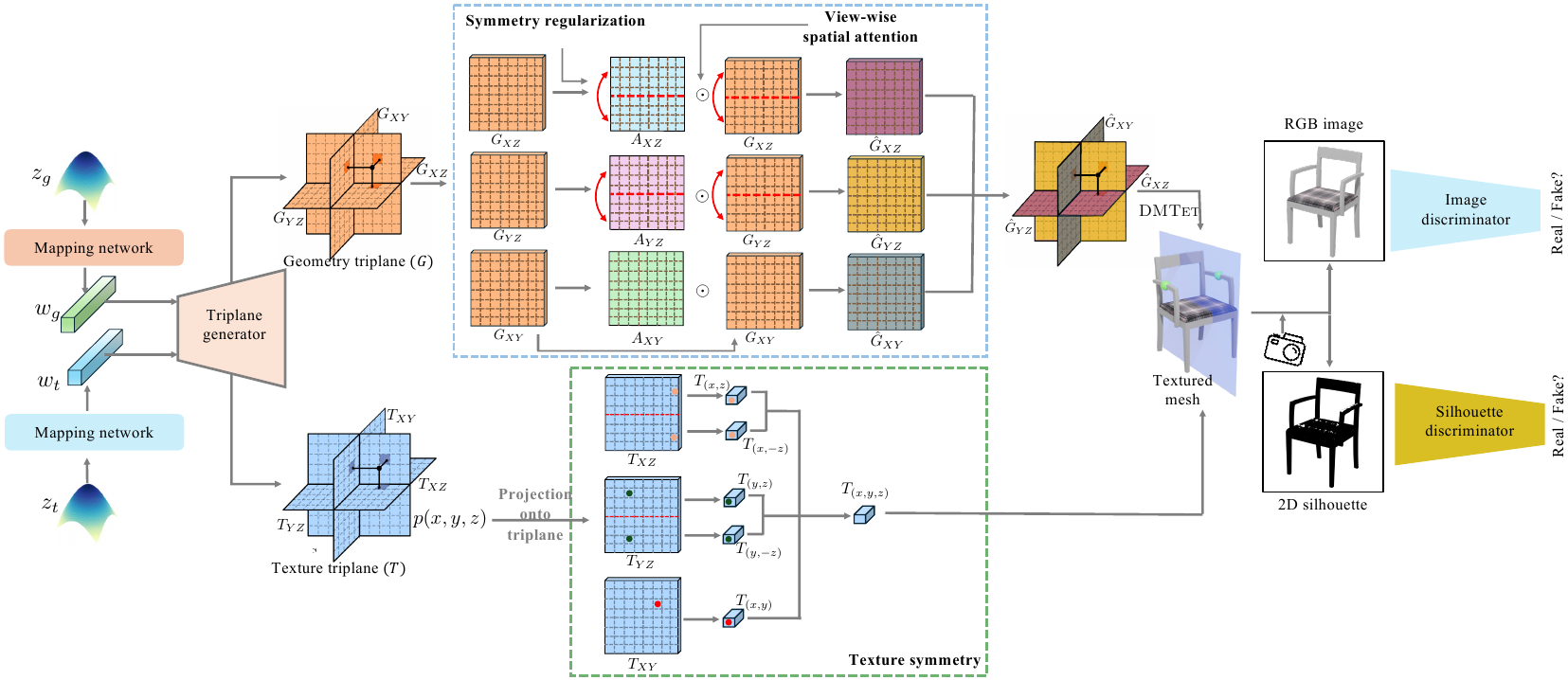}
    \caption{Overview of proposed \method{}. Random input vectors $z_{g}$ and $z_{t}$ are first mapped to a latent space ($w_g$ and $w_t$) and then fed into a shared generator to create the axis-aligned triplanes: geometry triplane $G$ and texture triplane $T$.
    We assume that the shapes being modeled have a symmetry plane ($XY$) such that a subset of the axis-aligned planes ($YZ,XZ$) can be regularized to exploit such symmetry. 
    We apply view-wise attention (Section~\ref{sec:sve}) on geometry triplane, and regulate both geometry triplane and attention map with reflectional symmetry (Section~\ref{sec:sr}). 
    We use DMTet method~\cite{dmtet} to extract a 3D mesh. 
    We describe a surface point $p$ with both the original and its reflective feature in texture triplane. 
    Using differentiable rendering~\cite{laine2020modular}, we render RGB images and their silhouettes from different camera angles. We then use two discriminators to determine whether the RGB and silhouette images are real or fake, without requiring the camera pose of real images.}
    \label{fig:sym3d}
    \vspace{-6mm}
\end{figure*}

%% file: sec/3_method.tex
\section{Method} \label{method}
In this section, we introduce the pipeline of our proposed 3D symmetry-aware textured mesh generation method (\method{}).
Section~\ref{sec:triplane} explains our triplane representation of 3D assets.
Section~\ref{sec:sve} and Section~\ref{sec:sr} describe the view-wise spatial attention and reflectional symmetry regularization.
Finally, Section~\ref{sec:loss} discusses the training process. 
An overview of the pipeline is illustrated in Figure \ref{fig:sym3d}.

\subsection{Triplane Representation of 3D Assets} \label{sec:triplane}
Our representation is built upon GET3D~\cite{get3d} for its capability to produce high-quality textured meshes and its effective separation of geometry from texture through triplanes~\cite{eg3d}.
We create two sets of triplanes, $G=\{G_{XY},G_{XZ},G_{YZ}\}$ for geometry and $T=\{T_{XY},T_{XZ},T_{YZ}\}$ for texture, effectively storing the shape and texture information of 3D objects.
A triplane consists of three axis-aligned orthogonal feature planes, each with size $N\times N \times C$ where $N$ is the spatial resolution and $C$ is the number of channels.
To create a 3D asset, we incorporate DMTet~\cite{dmtet} in the generator, which represents geometry as a signed distance field (SDF) defined on a deformable tetrahedral volume grid.
For any vertex $p = (x,y,z)$ in the tetrahedral grid, we calculate its geometry feature by first projecting it onto $XY$, $XZ$ and $YZ$ planes based on its 2D coordinate $(x,y),(x,z),(y,z)$, and then querying and aggregating features from these projections as described by: 
\begin{equation} 
    G_{(x,y,z)} = G_{(x,y)}+G_{(x,z)}+G_{(y,z)}.
    \label{Eq:tri_fea}
\end{equation}
The feature vector $G_{(x,y,z)}$ then represents the geometric features of $p$ and is used to infer the SDF value and deformation of the tetrahedral volume grid.
After computing SDF values and deformations, the differential marching tetrahedral algorithm extracts the explicit mesh.
To shade a surface point, the texture feature $T_{(x,y,z)}$ is calculated through a comparable process to Eq \ref{Eq:tri_fea} and is used to predict the RGB color.
With a known mesh structure, it simplifies computations by only requiring surface point queries, significantly reducing computational complexity.

To produce high-quality textured meshes, GET3D was trained with synthesized multi-view images rendered from various objects with camera pose annotations.
Our initial findings suggest that the geometry-based triplane representation, which uses three axis-aligned planes intended to capture the top, bottom, and side views of an object, often results in planes that are highly similar. The difficulty in training a triplane to have factorization features along axes in a 3D GAN setup arises during the optimization phase. 
Here, the generator focuses solely on creating a realistic 2D image from a specific view, overlooking the creation of a consistent high-quality 3D shape.
This issue becomes more noticeable when trained with single images, as the absence of multi-view information for each object complicates achieving a clear representation (see Figure \ref{fig:triplane_comp}).

To this end, we suggest adding view-wise spatial attention and symmetry regularization to the triplane model. 
Incorporating reflectional symmetry encourages the feature planes to respect the natural symmetry~\cite{mitra2013symmetry} of the object being modeled, which in turn assists the spatial attention in focusing each plane on unique parts of the scene, leading to a better overall 3D representation. 
By doing so, triplane learning will focus on both deceiving the discriminator and maintaining a holistic view of an object, enhancing the effectiveness and realism of the generated 3D models.

\subsection{View-wise Spatial Attention} \label{sec:sve}

In order to make each set of features robust and well-distributed over the three views, we develop view-wise spatial attention (VSA) for each of the three planes (Figure \ref{fig:SVA}). 

\input{fig/4_SVA}

The concept of attention in this context is well-documented in prior studies~\cite{senet,cbam}. 
Given that our planes are naturally divided into three distinct groups, we concentrate on learning attention that is specific to each plane, rather than applying the same attention across all planes. 
This view-specific attention tailors the focus to the unique aspects of each plane. For example, in Figure \ref{fig:SVA}, our aim is to ensure the $XY$ plane to capture an object's side view.

For each plane, this is achieved by initially aggregating the plane feature set along the channel axis via using two pooling (max, average) operations, generating two 2D maps. 
Subsequently, a localized convolutional layer is employed to derive the attention on the concatenation of two maps, which is then mapped to a weight value in the interval $[0,1]$ via a Sigmoid function:
\vspace{-2mm}
\begin{equation}
A_{XY} = \sigma \left( \text{Conv} \left( \text{mean}(\hat{G}_{XY}) \oplus \text{max}(\hat{G}_{XY}) \right) \right).
\label{Eq:attention_matrix}
\end{equation}
\vspace{-4mm}

To produce the attention-aware feature planes, we apply element-wise multiplication between the attention map and the original feature plane:
\begin{equation}
    \hat{G}_{XY} = A_{XY} \odot G_{XY}.
\label{Eq:attention_plane}
\end{equation}
We apply VSR to each feature plane within $G$, and finally we get a new set of triplanes:
\begin{equation}
    \hat{G} = \{ \hat{G}_{XY},\hat{G}_{YZ},\hat{G}_{XZ} \}.
    \label{Eq:attention_triplane}
\end{equation}

\vspace{-4mm}
\subsection{Reflectional Symmetry Regularization} \label{sec:sr}
\vspace{-2mm}
Reflectional symmetry is a common trait in many object categories\cite{mitra2013symmetry}, including those under our study. 
To leverage this characteristic, we impose a reflection-symmetry regularization on the geometry triplane $G$. 
Consider a chair as an example, in its triplane representation, we impose reflectional symmetry along the $XY$-plane. 
We propose two variants to enforce such symmetry:

\noindent {\bf Feature symmetry}: For the plane features
$G_{YZ}$ and $G_{XZ}$, which correspond to the front and down views, respectively, we enforce reflectional symmetry:
\begin{equation}
\mathcal{R}(G) = \| G_{YZ} -\textit{flip} (G_{YZ}) \|^2 + \| G_{XZ} - \textit{flip} (G_{XZ}) \|^2.
\label{Eq:symmetry_fea}
\end{equation}
\noindent {\bf Attention symmetry}: For the plane attention
$A_{YZ}$ and $A_{XZ}$, again corresponding to the front and bottom views, we enforce reflectional symmetry as follows:
\begin{equation}
\mathcal{R}(A)  = \| A_{YZ} -\textit{flip} (A_{YZ}) \|^2+ \|A_{XZ} - \textit{flip} (A_{XZ})\|^2.
\label{Eq:symmetry_att}
\end{equation}

While geometric symmetry is more naturally applicable to some categories, texture symmetry can also be beneficial \cite{symmnerf}. We implement a methodology where each pixel's features are combined with those of its symmetrical counterpart: 
\begin{equation}
    \mathcal{R}(T_{(x,y,z)}) = T_{(x,y)}+\frac{T_{(y,z)}+T_{(y,-z)}}{2}+\frac{T_{(x,z)}+T_{(x,-z)}}{2}.
    \label{Eq:symmetry_app}
\end{equation}
This technique is designed to efficiently capture and represent the object's features. 
The application of symmetry differs from its use in geometry, as it specifically focuses on the vertices of the generated mesh.
The use of symmetry prior enforces that the symmetric structure of objects is incorporated into the model's representation. 
This integration improves the representation ability of the triplane.

\input{tab/tab_chair_car}

\subsection{Training Objectives} \label{sec:loss}
We follow GET3D~\cite{get3d} and use the standard adversarial loss for training following:
\begin{equation}\label{Eq:GAN}
\begin{aligned}
L_{D} &= -\mathbb{E} [\log(1-\mathrm{D}(I_f))] - \mathbb{E}[\log(D(I_r))] \\
& + \lambda \mathbb{E}\left[\|\nabla_{I_r} D(I_r)\|^2\right],
 \\
L_{G} &= -\mathbb{E}[\log(D(I_f))]+\alpha \mathcal{R}(G)+\beta \mathcal{R}(A),
\end{aligned}
\end{equation}
where $I_{r}$ represents real data while $I_f$ denotes data produced by the generator (\ie RGB image, silhouettes). The third element in Eq \ref{Eq:GAN} is the gradient penalty, with $\lambda$ as the weighting coefficient for this term. The hyperparameters $\alpha$ and $\beta$ control the symmetry constraint on the geometry triplane and the attention maps, respectively. 
We have also adopted the shape regularization utilized in GET3D, which helps eliminate internal floating faces. 
Our discriminator does not require the camera pose of real images, which is inaccessible in most real-world datasets. Overall, by combining symmetry prior with other losses, our method produces realistic images, not just strictly symmetrical ones, making it suitable for a wide range of categories.


%% file: fig/4_SVA.tex
\begin{figure}[ht]
    \centering
\includegraphics[width=0.98\linewidth]{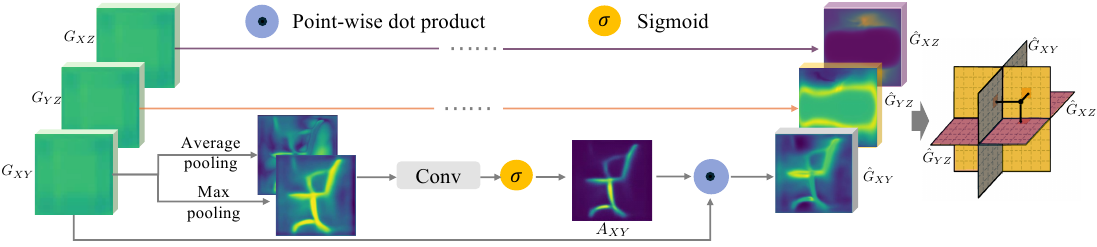}
    \caption{Illustration of proposed view-wise spatial attention (VSA) module. This module analyzes each plane individually, utilizing spatial features as guidance for attention. }
    \label{fig:SVA}
    \vspace{-4mm}
\end{figure}



%% file: tab/tab_chair_car.tex
\begin{table*}[t]
\begin{minipage}{0.48\textwidth}
\centering
\renewcommand{\arraystretch}{1.3}
\begin{adjustbox}{width=\linewidth}
\begin{tabular}{@{}clccccc@{}}
\hline
\multirow{2}{*}{Dataset} & \multicolumn{1}{c}{\multirow{2}{*}{Method}} & \multicolumn{2}{c}{COV (\%, $\uparrow$)} & \multicolumn{2}{c}{MMD ($\downarrow$)} & FID ($\downarrow$) \\ \cmidrule(l){3-7} 
    & \multicolumn{1}{c}{} & LFD & CD & LFD & CD & 3D \\ \midrule
    \multirow{3}{*}{Chair-S1} 
    & OP3D \cite{orthoplanes} &24.4	&0.09	&4183	&34.05&60.82\\
    & GET3D \cite{get3d} &\bf{66.31} & 15.03& 3412 & 14.98 & 55.17 \\
    & \method{} &64.58&\bf{58.53}&\bf{3227}&\bf{4.42}&\bf{38.34} \\
    \hline
    \multirow{3}{*}{Chair-S2} 
    & OP3D \cite{orthoplanes}&32.18&0.18&4764&11.8&74.83\\
    & GET3D \cite{get3d}&\bf{67.29} &15.38 &3754 &14.92 &63.35 \\
    & \method{} &67.02 &\bf{56.32} &\bf{3563} &\bf{4.74} &\bf{51.18}\\
    \hline
    \multirow{3}{*}{Chair-S3} 
    & OP3D \cite{orthoplanes}&27.76	&0.09	&4853	&15.50&76.40\\
    & GET3D \cite{get3d}&\bf{63.00} &16.00 &3837 & 15.52 &66.51\\
    & \method{} &62.86 &\bf{53.14} &\bf{3661} &\bf{4.94} &\bf{56.59}\\
\hline
\end{tabular}
\end{adjustbox}
\end{minipage}
\quad
\begin{minipage}{0.48\textwidth}
\centering
\renewcommand{\arraystretch}{1.3}
\begin{adjustbox}{width=\linewidth}
\begin{tabular}{@{}clccccc@{}}
\hline
\multirow{2}{*}{Dataset} & \multicolumn{1}{c}{\multirow{2}{*}{Method}} & \multicolumn{2}{c}{COV (\%, $\uparrow$)} & \multicolumn{2}{c}{MMD ($\downarrow$)} & FID ($\downarrow$) \\ \cmidrule(l){3-7} 
    & \multicolumn{1}{c}{} & LFD & CD & LFD & CD & 3D \\ \midrule
    \multirow{3}{*}{Car-S1} 
    &OP3D \cite{orthoplanes} &29.24&0.07&2593&13.48&34.23\\
    & GET3D \cite{get3d} &58.30&28.25&1461&1.42&29.69\\
    & \method{}&\bf{63.35}&\bf{36.13}&\bf{1284}&\bf{1.21} &\bf{23.07}\\
    \hline
    \multirow{3}{*}{Car-S2} 
    & OP3D \cite{orthoplanes}&15.69&0.07&3404&18.5&41.49\\
    & GET3D \cite{get3d} &55.96	&19.88	&1731	&1.77&34.60\\
    & \method{} &\bf{65.46}	&\bf{32.86}	&\bf{1708}	&\bf{1.50} & \bf{31.35}\\
    \hline
    \multirow{3}{*}{Car-S3} 
    &OP3D \cite{orthoplanes}&12.43&0.13&3737&20.67&48.19\\
    & GET3D \cite{get3d} &47.79&15.86&1747&1.80&36.39\\
    & \method{} &\bf{50.13}	&\bf{19.88}	&\bf{1709} &\bf{1.69} &\bf{32.81}\\
\hline
\end{tabular}

\end{adjustbox}
\end{minipage}
\caption{Quantitative results on ShapeNet-Chair and ShapeNet-Car. The best result is shown in \textbf{bold}, and MMD-CD scores are multiplied by $10^3$. 
}
\label{tab:chair_car}
\vspace{-6mm}
\end{table*}

%% file: sec/4_exp.tex
\section{Experiment} \label{exp} 
\subsection{Settings} \label{datasets}
\noindent {\bf Datasets.}
We conduct experiments on the synthetic ShapeNet~\cite{shapenet} dataset and the real-world Amazon Berkeley Objects (ABO)~\cite{collins2022abo} dataset.
Following GET3D~\cite{get3d}, we focus on car and chair categories from ShapeNet, where each object is represented by 24 different views. 
As suggested by~\cite{get3d}, we split each category into training (70\%), validation (10 \%), and testing (20\%) sets. 
To simulate a more realistic training scenario similar to natural images, our approach only uses a \emph{single view} of each object in the training set.
We define three scenarios based on the range of azimuth angles for the selected view: 
Scenario 1 (S1) spans 0-360 degrees, Scenario 2 (S2) covers 0-180 degrees, and Scenario 3 (S3) encompasses 0-120 degrees.
For S2 and S3, we adopt random view flips to enhance view diversity.
For ABO datasets, we run experiments on its chair category, which has 1158 objects.
All experiments are conducted at a resolution of $1024 \times 1024$.

\noindent {\bf Competitors.}
To the best of our understanding, \method{} marks the initial attempt to integrate a symmetry prior into a triplane representation. While, GET3D is unique in its approach to separate geometry and texture during the training of a 3D generative model, making it the benchmark in this field. As such, GET3D serves as our primary reference point. We also draw comparisons with another SOTA work in 3D-aware image synthesis, OrthoPlanes for 3D (OP3D)~\cite{orthoplanes}, which enhances the triplane representation by maintaining information related to the projection distance. 

\input{fig/4_RGB_compare}
\input{fig/6_cam}
\input{fig/2_shape_texture_compare}

\noindent {\bf Implementations.}
In real-world settings, accurately determining camera poses can be difficult. As a result, we adopt a strategy of training the discriminator without camera pose condition, opting instead for a fixed camera distribution approach, as demonstrated to be effective in previous studies \cite{schwarz2020graf,get3d}. 
For all our experiments, we utilize the camera distribution defined in GET3D. It's important to mention that the camera distribution is not completely covered in S3.
For hyper-parameters, we set $\alpha=100$, $\beta=10$ in the experiments.
Additionally, we adopt the same setup of~\cite{get3d} including the training configuration as given in its open source code\footnote{\url{https://github.com/nv-tlabs/GET3D}}. Experiments are done on 8 A100 GPUS.

\noindent {\bf Metrics.}
To assess the quality of the generated objects, we examine both their geometry and texture. 
For geometric, we use Coverage (COV) and Minimum Matching Distance (MMD)~\cite{achlioptas2018learning} metrics, which assess the average quality and diversity of the shapes. The comparison between two shapes is using the Chamfer Distance (CD) and the Light Field Distance (LFD)~\cite{chen2017distance}, measuring their similarity. For OP3D \cite{orthoplanes}, we use marching cubes to extract the underlying geometry. For texture, we utilize the Fréchet Inception Distance (FID) \cite{heusel2017gans}, which is applied to 2D rendered images of the objects.

\vspace{-2mm}
\subsection{Main Results}
\vspace{-2mm}
\noindent {\bf Quantitative Results.} The quantitative results for the ShapeNet chairs and cars are presented in Table~\ref{tab:chair_car}. 
Our observations are as follows: 
{\bf(1)} In comparison to OP3D \cite{orthoplanes}, which utilizes a single triplane representation to encode both shape and texture, GET3D \cite{get3d} and our \method{}, which separate shape and texture learning, exhibit improved generation in quality and diversity. 
{\bf(2)} Models trained on various splits show that S1 outperforms S2, which in turn outperforms S3. Specifically, S1 encompasses a full 360-degree view of an object category, S2 achieves a 360-degree view through image flipping, while S3 is limited to a 240-degree view. The results further suggest that incorporating views from various angles, including those from other objects, enhances the learning of a 3D representation.
{\bf(3)} All models are trained without conditioning discriminator on camera poses. Our approach, which incorporates symmetry assumed in a canonical space, yields superior results in both COV and MMD metrics. This indicates that the shapes learned by \method{} closely resemble those in a canonicalized form.
{\bf(4)} {\method{}} outperforms GET3D in majority metrics, demonstrating the benefits of integrating symmetry regularization and view-wise spatial attention into 3D representations. Consequently, {\method{}} proves to be an effective approach that enhances 3D representation. 

\noindent {\bf Qualitative Comparisons.}
Figure~\ref{fig:RBG_compare} provides qualitative comparisons against competitors in terms of generated 2D images. In general, \method{} achieves a more realistic appearance across different settings. In the more challenging settings S2, S3, and real-world ABO-chair, OP3D struggles to generate a complete scene, and GET3D tends to generate distortions.
Figure~\ref{fig:cam_compare} provides comparisons across various camera positions. GET3D and \method{} are able to generate view-consistent images across different camera views but OP3D fails to do that. This demonstrates that decoupling shape and texture benefits the 3D-aware generation.

Since both GET3D and \method{} generate textured meshes, we export their shapes into Blender and show their comparison in Figure~\ref{fig:shape_texture_compare}. \method{} significantly outperforms GET3D in creating textured meshes. It consistently delivers more regular and higher fidelity representations across various objects. 
While GET3D struggles with generating a complete shape, often producing armchairs with missing halves, uneven chair backs, broken carriage, and other irregularities. In contrast, our method provides uniform and complete shapes. 
This advantage becomes even more apparent when working with incomplete views (S3); GET3D is prone to creating fragmented shapes due to the absence of certain viewpoints. \method{} shows that with symmetry as a structural prior, the generator can learn to produce complete and accurate shapes, even when trained on datasets with limited views. Additionally, when applied to real-world datasets, the shortcomings of GET3D become evident through the creation of chairs with unrealistic features, such as five legs or irregularly shaped supports, highlighting the effectiveness of our method.  

\subsection{Properties of Learned Triplane}


\noindent{\bf View-wise Triplanes.}
We use a similarity matrix to measure difference among feature maps in the geometry triplane. The geometry triplane is a $N \times N \times 3C$ feature tensor, which we flatten to a 2D tensor $3C \times N^2$.
\input{fig/7_triplane}
Subsequently, we compute the similarity matrix for this 2D tensor. 
Each entry within this matrix is a cosine similarity between the two channels. 
The rendered 2D images, similarity matrix, along with 3 selected feature maps from $XY$, $YZ$, and $XZ$ planes from both GET3D (\ie $G_{XY}$, $G_{YZ}$, $G_{XZ}$) and \method{} ($\hat{G}_{XY}$,$\hat{G}_{YZ}$,$\hat{G}_{XZ}$), are visualized in Figure \ref{fig:triplane_comp}. This comparison reveals that our method produces planes more specific to each view, demonstrating increased similarity within each plane but high discrimination across different planes.
Furthermore, the $XY$ plane in our method clearly exhibits a side view pattern, offering a more distinct representation. Although the $YZ$, $XZ$ planes do not capture front and down views in a precise way, they still adhere to the symmetry assumption. We argue that the enforced symmetry regularization plays a crucial role in driving a clearer representation in the $XY$ plane.


\noindent{\bf Robustness to Biased Views.}
Figure~\ref{fig:chair_s3} shows the results from models trained on the incomplete view dataset: ShapNet Chair-S3. Overall, \method{} shows its superiority in generating complete objects.

\input{fig/9_orientation_ab}
\noindent{\bf Consistent Camera Orientations.}
The fixed camera distribution strategy ensures that all objects produced by both GET3D and \method{} are uniformly positioned and oriented. 
By applying symmetry regularization, our method secures a consistent orientation for objects in the chair category with respect to the symmetry plane (Figure~\ref{fig:oritentation_comp}).

\subsection{Further Analysis}
\noindent {\bf The effect of different components.}
We validate the design of our framework by ablating four components using the ShapeNet Chair-S1 dataset.
SVE introduces view-wise spatial enhancement in Section \ref{sec:sve}, $\mathcal{R}(G)$ adds symmetry in feature maps as in Eq.~\ref{Eq:symmetry_fea}, $\mathcal{R}(A)$  adds symmetry in attention maps as in Eq.~\ref{Eq:symmetry_att}, $\mathcal{R}(T)$ supplements a point's feature with its symmetric counterpart as in Eq.~\ref{Eq:symmetry_app}. Our findings indicate that each component independently improves the quality of generation, with the collective implementation of all elements resulting in the most effective model.

\input{tab/tab_ab}

\noindent {\bf With window self-attention.}
To show the effectiveness of the proposed attention module, we compare \method{} against a variant that uses window self-attention \cite{ramachandran2019stand} across three views, without employing attention symmetry. 
Considering memory limitations, we chose two window sizes: Wins=4 and Wins=8. Table \ref{tab:winsa} indicates that SYM3D, while adding negligible increases in parameters and computational cost, results in significant improvements.
\input{tab/tab_wins}

\noindent {\bf On a more complex category.}
For evaluating the generality of proposed method on more complex categories, we test on airplane dataset in ShapeNet~\cite{shapenet}. We render 24 views for each object and select one view as training set. As shown in Table \ref{tab:plane}, we obtain consistent results as on ShapeNet-Chair and ShapeNet-Car in Table \ref{tab:chair_car}. 
\input{tab/tab_plane}


\noindent {\bf Applied on the text-to-3D task.}
Current text-to-3D task usually utilize text-to-image generation models as guidance. The training process distills one image at a time, facing challenges in maintaining consistent views and precise geometry, leading to texture misalignments, asymmetry, incoherent appearances, or severe ``Janus effect" issues where features like faces or eyes appear repeatedly and unnaturally \cite{bensadoun2024meta} in generated object.  
Symmetry assumption encodes information from unseen views, providing a global constraint that defines a canonical frame, crucial for accurate model alignment and generation.

To demonstrate the generalizability of our method, we have applied our proposed symmetry regularization to this task. 
MTN~\cite{yi2023progressive} introduces a multi-scale triplane representation for the text-to-3D task. 
We apply symmetric regularization (see Eq~\ref{Eq:symmetry_fea}) to these multi-scale triplanes. We develop two versions: W/O SYM and W SYM, whose results are in Figure~\ref{fig:text23d}. Our observations show that W/O SYM often generates salient artifacts, such as extra or malformed legs and ears when modeling a cat, and a dog with three forelegs. Conversely, W SYM produces cats and dogs without these defects, proving that symmetric priors effectively eliminate such artifacts in 3D model creation.
\input{fig/10_text23d}




%% file: fig/4_RGB_compare.tex

\begin{figure*}[t]
\centering
\includegraphics[width=0.88\textwidth]{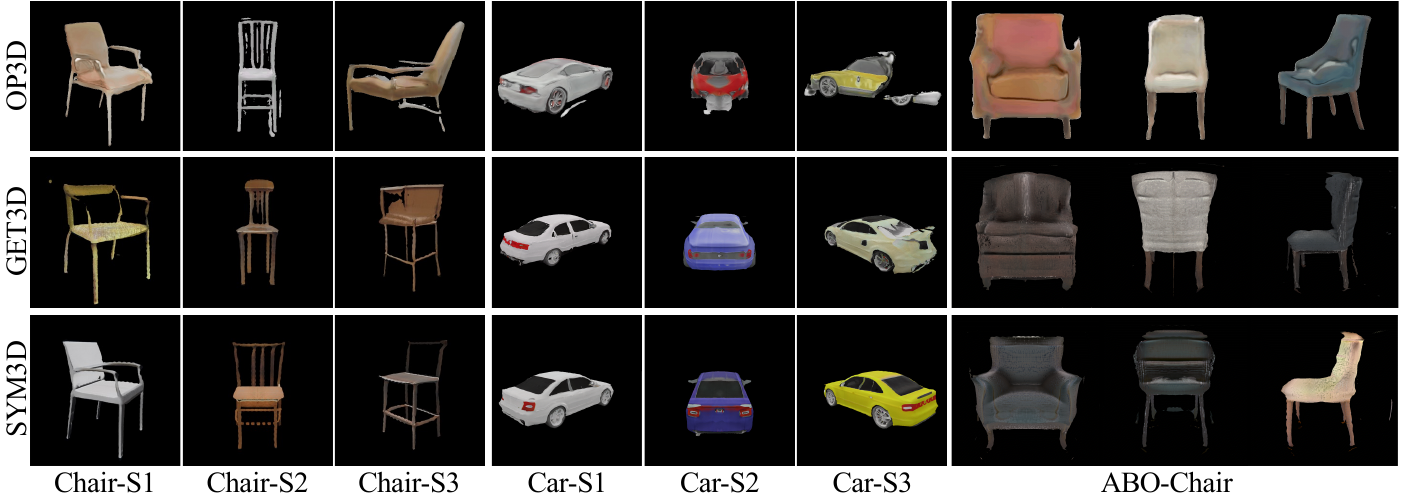}
\vspace{-4mm}
\caption{Qualitative comparison of \method{} against OP3D and GET3D on generated images. \method{} produces images with sharp details and a high diversity of shapes.}
\label{fig:RBG_compare}
\vspace{-2mm}
\end{figure*}

%% file: fig/6_cam.tex
\begin{figure*}[h]
\centering
\begin{subfigure}{0.443344\linewidth}
    \includegraphics[width=\linewidth]{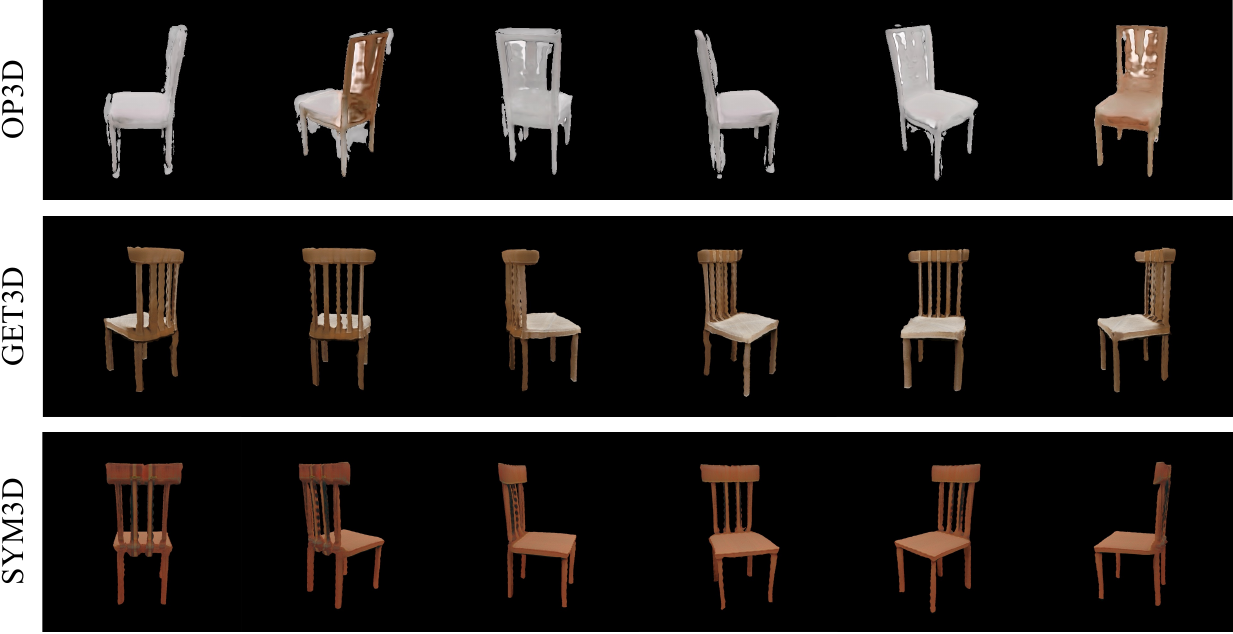}
    \caption{Trained on ShapeNet Chair-S1}
\end{subfigure}
\begin{subfigure}{0.4224\linewidth}
    \includegraphics[width=\linewidth]{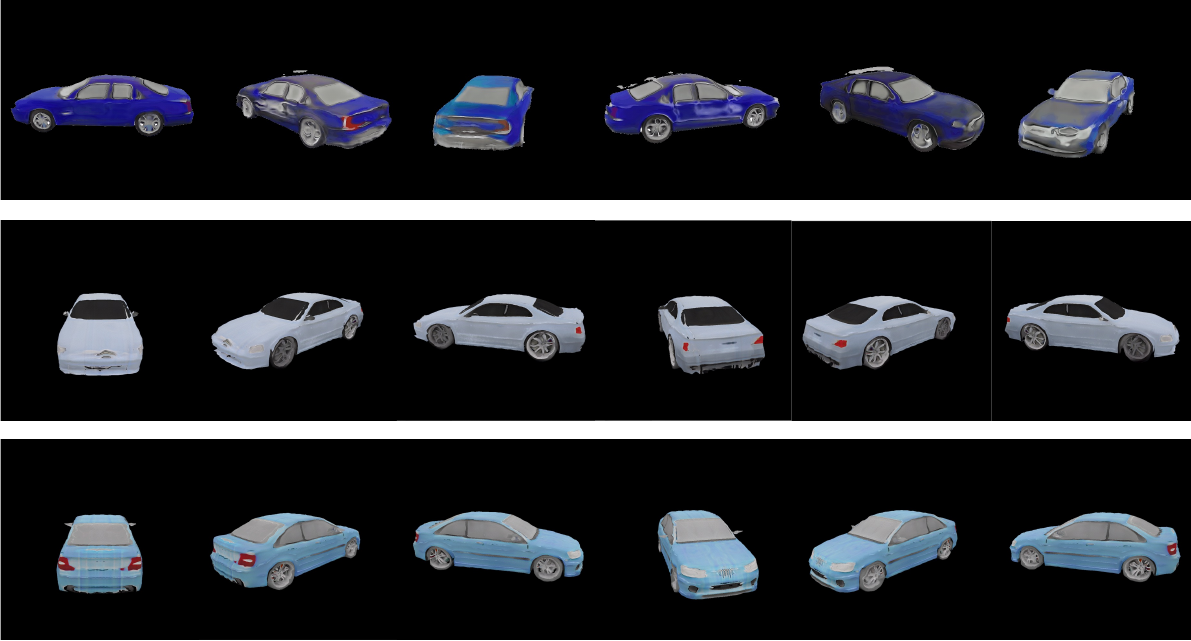}
    \caption{Trained on ShapeNet Car-S1}
\end{subfigure}  
\vspace{-4mm}
\caption{Rendered RGB images across different camera views.}
\label{fig:cam_compare}
\vspace{-4mm}
\end{figure*}

%% file: fig/2_shape_texture_compare.tex

\begin{figure*}
\centering
\includegraphics[width=0.96\textwidth]{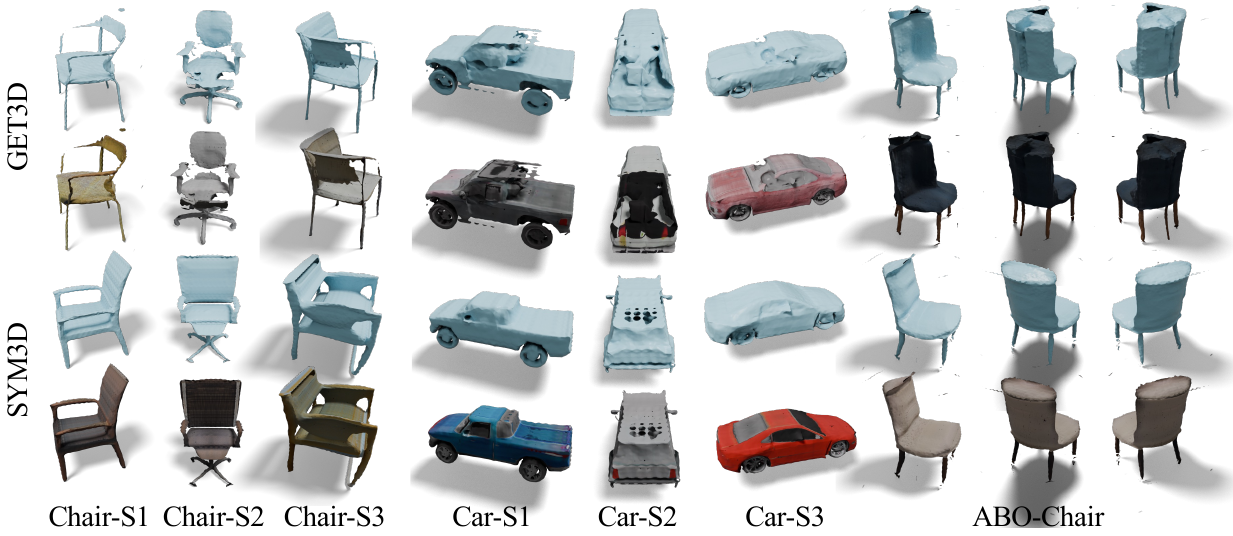}
\caption{Comparison on shapes generated by GET3D and \method{} rendered in Blender. ~\method{} significantly outperforms GET3D in creating textured meshes. It consistently delivers more regular and higher fidelity representations across various objects. In contrast, GET3D often produces armchairs missing halves, uneven chair back, broken carriage, and other irregularities.}
\label{fig:shape_texture_compare}
\vspace{-4mm}
\end{figure*}

%% file: fig/7_triplane.tex
\begin{figure}[t]
    \centering
    \includegraphics[width=0.88\linewidth]{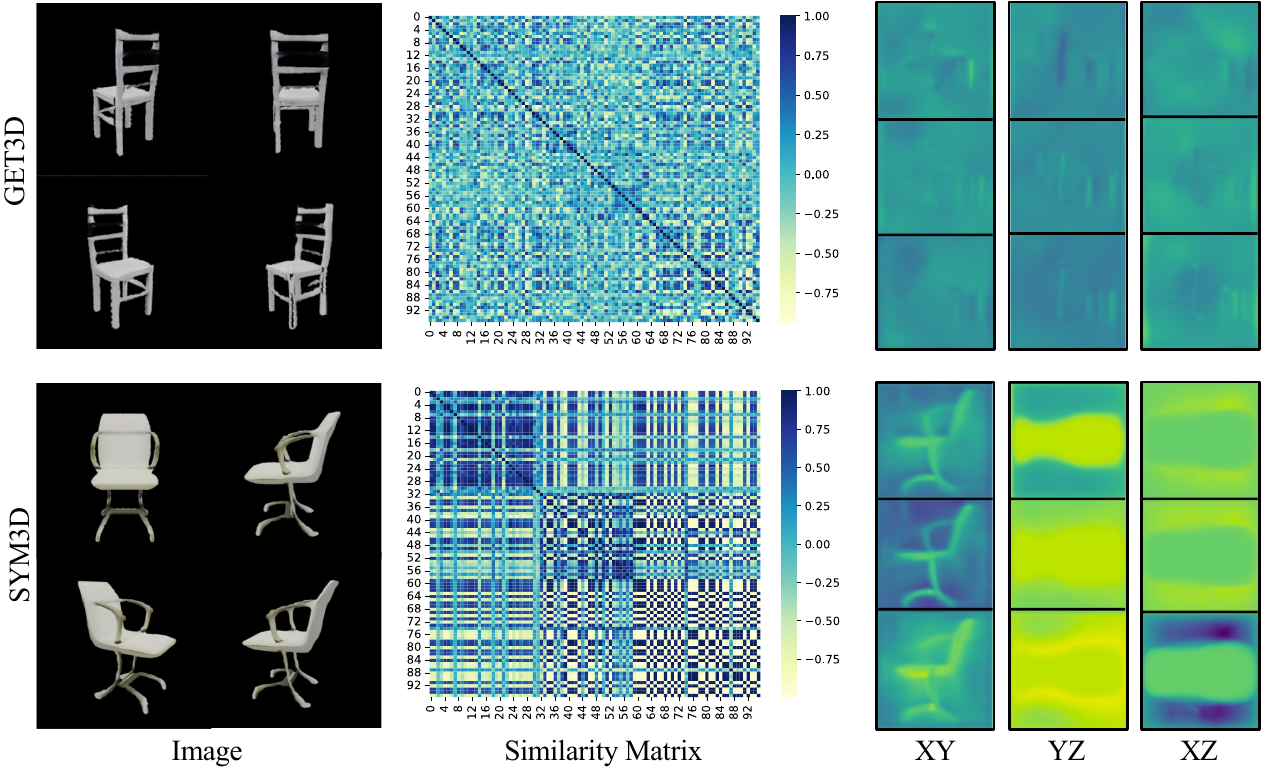}
    \vspace{-4mm}
    \caption{Geometry triplane comparison between GET3D and \method{}. From left to right: rendered 2D image, similarity matrix across channels, feature maps from each plane. We note that \method{} displays enhanced view-wise properties.}
    \label{fig:triplane_comp}
    \vspace{-6mm}
\end{figure}


%% file: fig/9_orientation_ab.tex
\begin{figure}
\centering
\includegraphics[width=0.88\linewidth]{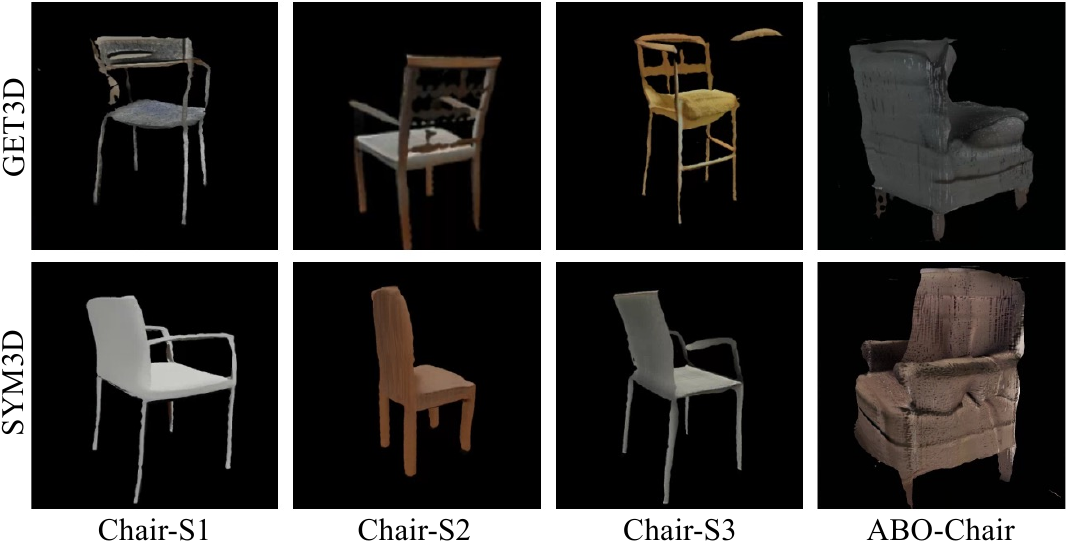}
\vspace{-3mm}
\caption{Images rendered from different models given a fixed camera view.}
\label{fig:oritentation_comp}

\vspace{-8mm} 
\end{figure}

%% file: tab/tab_ab.tex
\vspace{-2mm}
\begin{table}[h]
\centering
\begin{tabular}{c|c|c|c|c}
\hline
SVE     &$\mathcal{R}(G)$  &$\mathcal{R}(A)$  &Tex      &FID\\
\hline
\xmark  &\xmark    &\xmark    &\xmark   &55.17\\
\xmark  &\cmark    &\xmark    &\cmark   &47.87\\
\cmark  &\xmark    &\cmark    &\cmark   &43.43\\
\cmark  &\cmark    &\cmark    &\xmark   &41.09\\
\cmark  &\cmark    &\cmark    &\cmark   &38.23\\
\hline
\end{tabular}
\vspace{-2mm} 
\caption{ablation studies}
\vspace{-5mm} 
\label{tab:ab_loss}
\end{table}


%% file: tab/tab_wins.tex
\vspace{-2mm}
\begin{table}[h]
\centering
\begin{tabular}{c|c|c|c}
\hline
Method    &Added Params &Added Flops  &FID  \\
\hline
GET3D~\cite{get3d} &0 &0 &55.17\\
WinS=4&4616&1.0G&49.83\\
WinS=8&6024&1.6G&46.73\\
\method{}&294&19.3M&38.23 \\
\hline
\end{tabular}
\vspace{-2mm} 
\caption{Comparison using the self-attention module, tested on ShapeNet Chair-S1.}
\label{tab:winsa}
\vspace{-3mm} 
\end{table}

%% file: tab/tab_plane.tex
\begin{table}[h]
\centering
\begin{tabular}{@{}lccccc@{}}
\hline
\multicolumn{1}{c}{\multirow{2}{*}{Method}} & \multicolumn{2}{c}{COV (\%, $\uparrow$)} & \multicolumn{2}{c}{MMD ($\downarrow$)} & FID ($\downarrow$) \\ \cmidrule(l){2-6} 
& LFD & CD & LFD & CD & 3D \\ \midrule    
GET3D~\cite{get3d}&53.73&7.17&4980&5.37&40.66 \\
\method{} &\bf{57.85}&\bf{49.07}&\bf{4074}&\bf{1.15}&\bf{32.61}\\
\hline
\end{tabular}
\vspace{-2mm}
\caption{Generality on a more complex category. Experiments are done on ShapeNet Airplane.}
\label{tab:plane}
\vspace{-4mm}
\end{table}

%% file: fig/10_text23d.tex
\begin{figure}[ht]
\vspace{-2mm}
    \centering
    \includegraphics[width=0.88\linewidth]{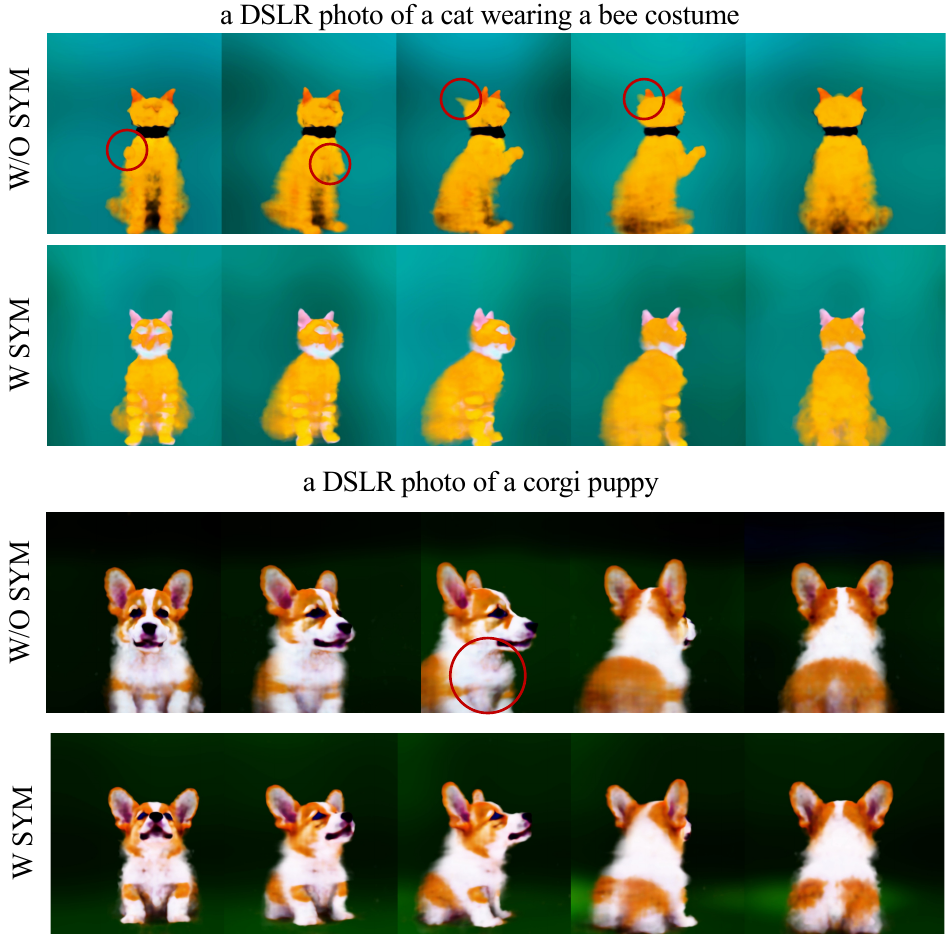}
        \vspace{-4mm}
    \caption{Comparisons on text-to-3D. W/O SYM produces artifacts in rendered images, while W SYM with symmetry regularization loss avoids this.}
    \label{fig:text23d}
    \vspace{-4mm}
\end{figure}

%% file: sec/5_conclusion.tex
\vspace{-4mm}
\section{Conclusion} \label{sec:conclusion}
\vspace{-2mm}
We present \method{} for learning symmetric triplanes for improving 3D-awareness of GANs when trained on single images without camera pose annotation. 
While effective it has several {\bf limitations} that future work should investigate. Global reflectional symmetry is not always satisfied and some objects satisfy other geometric transformations symmetry such as reflections, translations, rotations, or combinations in local parts (\ie car tires, chair legs). In future work, we plan to extend our approach to create a large real dataset of common object categories and combat the canonicalization issue as in~\cite{agaram2023canonical} and symmetry issue as in~\cite{seo2022reflection}.


